\documentclass[11pt]{article}
\usepackage{graphicx} 
\usepackage{amsmath}
\usepackage{lipsum}
\usepackage{pdfcomment}

\usepackage[table,xcdraw]{xcolor} 
\usepackage{subcaption}
\usepackage[utf8]{inputenc}
\usepackage{longtable}
\usepackage{graphicx}
\usepackage{float}
\usepackage{booktabs}
\usepackage{float} 
\usepackage{geometry}
\usepackage{lineno}
\usepackage{authblk}
\usepackage{tcolorbox}
\usepackage{rotating}
\usepackage{booktabs}
\usepackage{threeparttable}
\usepackage{geometry} 
\usepackage{lscape}
\usepackage{siunitx}  
\usepackage{adjustbox} 
\usepackage{pgfplots}
\usepackage{caption}
\usepgfplotslibrary{groupplots}

\usepackage[numbers,sort&compress]{natbib}

\usepackage{hyperref}
\hypersetup{
  colorlinks=true,
  linkcolor=blue,       
  citecolor=blue,       
  urlcolor=blue         
}

\usepackage{cite}

\usepackage{array}
\usepackage{multirow}

\usepackage{amssymb}
\usepackage{pifont}

\usepackage[colorinlistoftodos]{todonotes}

\usepackage{wrapfig}

\usepackage{cite}

\usepackage{amsthm}

\usepackage{titlesec}
\usepackage{titletoc}
\titleclass{\subsubsubsection}{straight}[\subsection]
\newcounter{subsubsubsection}[subsubsection]
\renewcommand\thesubsubsubsection{\thesubsubsection.\arabic{subsubsubsection}}
\titleformat{\subsubsubsection}{\normalfont\normalsize\bfseries}{\thesubsubsubsection}{1em}{}
\titlespacing*{\subsubsubsection}{0pt}{3.25ex plus 1ex minus .2ex}{1.5ex plus .2ex}
\titleformat*{\section}{\fontsize{11}{12}\selectfont\bfseries}
\titleformat*{\subsection}{\fontsize{11}{12}\selectfont\bfseries}
\titleformat*{\subsubsection}{\fontsize{11}{12}\selectfont\bfseries}
\titleformat*{\paragraph}{\fontsize{11}{12}\selectfont\bfseries}
\titleformat*{\subparagraph}{\fontsize{11}{12}\selectfont\bfseries}

\titlecontents{subsubsubsection}
[8em] 
{\small}
{\hyperlink{toc.\thecontentslabel}{\thecontentslabel.} }
{}
{\ \titlerule*[.5pc]{.}\contentspage}

\titlespacing*{\subsubsubsection}{0pt}{3.25ex plus 1ex minus .2ex}{1.5ex plus .2ex}

\usepackage{caption}
\usepackage{microtype}

\usepackage{enumitem}
\setlist{itemsep=0em}

\usepackage{fancyhdr} 
\usepackage{lastpage} 

\usepackage{listings}
\usepackage{xcolor}

\definecolor{codegreen}{rgb}{0,0.6,0}
\definecolor{codegray}{rgb}{0.5,0.5,0.5}
\definecolor{codepurple}{rgb}{0.58,0,0.82}
\definecolor{backcolour}{rgb}{0.95,0.95,0.92}

\lstdefinestyle{mystyle}{
    backgroundcolor=\color{backcolour},   
    commentstyle=\color{codegreen},
    keywordstyle=\color{magenta},
    numberstyle=\tiny\color{codegray},
    stringstyle=\color{codepurple},
    basicstyle=\ttfamily\footnotesize,
    breakatwhitespace=false,         
    breaklines=true,                 
    captionpos=b,                    
    keepspaces=true,                 
    numbers=left,                    
    numbersep=5pt,                  
    showspaces=false,                
    showstringspaces=false,
    showtabs=false,                  
    tabsize=2
}

\title{Considering Spatial Structure of the Road Network in Pavement Deterioration Modeling}

\author[1]{Lu Gao, Ph.D.}
\author[2]{Ke Yu, Ph.D.}
\author[3]{Pan Lu, Ph.D.}

\affil[1]{Department of Civil and Environmental Engineering, University of Houston\\
\texttt{lgao5@central.uh.edu}}
\affil[2]{School of Computing and Information, University of Pittsburgh \\
\texttt{gatechke@gmail.com}}
\affil[3]{College of Business, North Dakota State University \\
\texttt{pan.lu@ndsu.edu}}
\date{}

\begin{document}

\maketitle

\section*{Abstract}
Pavement deterioration modeling is important in providing information regarding the future state of the road network and in determining the needs of preventive maintenance or rehabilitation treatments. This research incorporated spatial dependence of road network into pavement deterioration modeling through a graph neural network (GNN). The key motivation of using a GNN for pavement performance modeling is the ability to easily and directly exploit the rich structural information in the network.  This paper explored if considering spatial structure of the road network will improve the prediction performance of the deterioration models. The data used in this research comprises a large pavement condition data set with more than a half million observations taken from the Pavement Management Information System (PMIS) maintained by the Texas Department of Transportation. The promising comparison results indicates that pavement deterioration prediction models perform better when spatial relationship is considered.

\noindent \textbf{Keywords}: Pavement Management, Deterioration Model, Graph Neural Network, Deep Learning, Pavement Performance

\section{Introduction}\label{introduction}

The spatial relationships between infrastructure facilities have been studied in the area of asset management by previous researchers \citep{zou2012incorporating,jahanbakhsh2016estimating,atef2014modeling,bernhardt2004agent,mcdaniels2007empirical,rahman2009identification,panzieri2004agent,dhatrak2020considering}. For example, \citet{zou2012incorporating} presented an approach to address pavement management decision problems at airports with multiple runways by considering functional dependence between runways. \citet{atef2014modeling} presented a framework to model spatially and functionally dependent assets. The developed model can be used to determine an asset's degree of connectivity with its neighbors. Another group of researchers have studied the economic dependence of road network structure. In those cases, often the cost of executing work on upholding many components’ integrity at the same time can be cheaper than doing the same work on individual components collectively. This includes the scenarios where doing any kind of maintenance work requires large amounts of setup and preparation work to be done in advance. \citet{bernhardt2004agent} noted that pavements are interconnected through geography, which implied the economies of scale in contracting long stretches of pavement for rehabilitation and the diseconomies of scale in terms of the disruption to users. Some other researchers have studied the failure dependence among infrastructure facilities or systems. For example, \citet{mcdaniels2007empirical} developed an analytical framework to characterize infrastructure failure interdependencies. The authors studied how extreme events lead to failures of other spatially-connected infrastructure systems, e.g., given a major electrical power outage. \citet{rahman2009identification} used public domain failure reports to identify the origin of these failures and their propagation patterns. The authors studied historical records to determine the causes of infrastructure failures and the impact of failures in spatial and temporal dimensions. \citet{panzieri2004agent} analyzed performance degradation induced by the spreading of failures in order to emphasize the most critical links existing among different critical infrastructure network. Spatial dependency has also been taken into consideration in developing pavement deterioration models through recognizing similarities between adjacent sections. This is usually conducted through developing separate models for pavement in different categories (climate, traffic, material). Despite these efforts by previous research, there is a lack of studies that take neighboring sections’ information into consideration when developing pavement deterioration models. It is possible that the prediction of a pavement section’s deterioration can be improved by taking into consideration neighboring sections’ condition information because they are all exposed to a set of common factors that make them fail, such as loading, operation or environmental factors.

Recently, with the development of big data and artificial intelligence, deep learning models have received considerable attention in the pavement performance modeling area \citep{deng2024short,gao2025modeling,alnaqbi2024machine,peng2025evaluating,marcelino2020transfer,gao2021deep,hosseini2020use,yu2023pavement,garcia2023integration,gao2022missing,haddad2022use,gao2021detection}. Compared with traditional models, deep learning models are designed to make more accurate prediction results. For example, \citet{lee2019development} developed a pavement deterioration prediction model based on deep neural network and Recurrent Neural Network (RNN) with LSTM circuits. They found that the performance and accuracy of the LSTM model was superior. \citet{choi2019development} predicted the deterioration of a road pavement by using monitoring data and a LSTM framework. The constructed algorithm predicting the pavement condition index for each section of the road network for one year by learning from the time series data for the preceding 10 years. \citet{hosseini2020use} developed deterioration models for Pavement Condition Index (PCI) as a function of time using two modeling approaches: deep learning model of LSTM and individual regression models. A comparison was made between the two approaches and the results show that the LSTM model achieved a higher prediction accuracy over time for all different pavement types. \citet{gao2023deep} employed a deep-learning based deterioration model through a CNN-LSTM combined framework to detect if an M\&R treatment was applied to a pavement section during a given time period. \citet{haddad2022use} used a deep neural network for pavement rutting prediction. The predictive capability of the proposed model was compared to a multivariate linear regression model fitted using the same dataset. It is found that the deep neural network rutting prediction model enhanced predictive power compared to commonly used models in the literature. \citet{zhou2021predicting} also applied the LSTM model to predict asphalt concrete (AC) pavement IRI, utilizing datasets extracted from the Long-Term Pavement Performance (LTPP) database. \citet{gao2022missing} introduced a convolutional graph neural network for imputing missing pavement condition data in pavement management systems, outperforming standard machine learning models.

In this research, we investigated applying a special class of deep learning methods called graph neural network (GNN) for pavement deterioration modeling. GNN is a class of deep learning methods designed to perform inference on data described by graphs. The objective of this research is to study if the prediction of a pavement section’s deterioration can be improved by taking neighboring sections’ condition information into consideration.

\section{Methodology}\label{methodology}

GNN is one of the fastest growing areas in deep learning. Its popularity lies in its strength to utilize the graph structure of data in network format such as transportation network, social network, and biology \citep{zhou2020graph}. Compared with traditional machine learning and deep learning models, the advantage of GNN is that it is able to utilize the spatial relationship between data points and aggregate information through graph edges. In the case of road network in this research, individual pavement sections are modelled as graph nodes, and the connection between the sections are modeled as graph edges. The way GNN works is to create node embeddings where information of individual node is represented as low dimensional vectors and the links (relationships between nodes) are maintained in the graph structure. With this setting, the model can be used for graph classification, node classification or regression.  

In this research, we used the Graph Sample and Aggregate (GraphSAGE) model \citep{hamilton2017inductive}, which is one of the common used GNN models. GraphSAGE allows training large-scale networks with mini-batch setting, where the model learns a function that outputs node embeddings based on the neighborhood of a node rather than learning all of the node embeddings directly. This significantly limits the memory and time needed to train the model for large networks.

    We define the road network as a graph $G(V,E)$ with $V$ representing nodes (pavement sections) and $E$ indicating the connections between them. 
The information (e.g., traffic, environment, pavement type) of a section $v$ is represented as the vector $\mathbf{x}_v$, and the complete set of node feature vectors as $\{\mathbf{x}_v \mid \forall v \in V\}$. 
The number of graph layers, $K$, in GraphSAGE specifies the number of hops the information of each node can travel across the graph.

\begin{figure}[htbp]
    \centering
      \includegraphics[width=0.9\linewidth,
                   trim=20pt 200pt 30pt 80pt, 
                   clip]{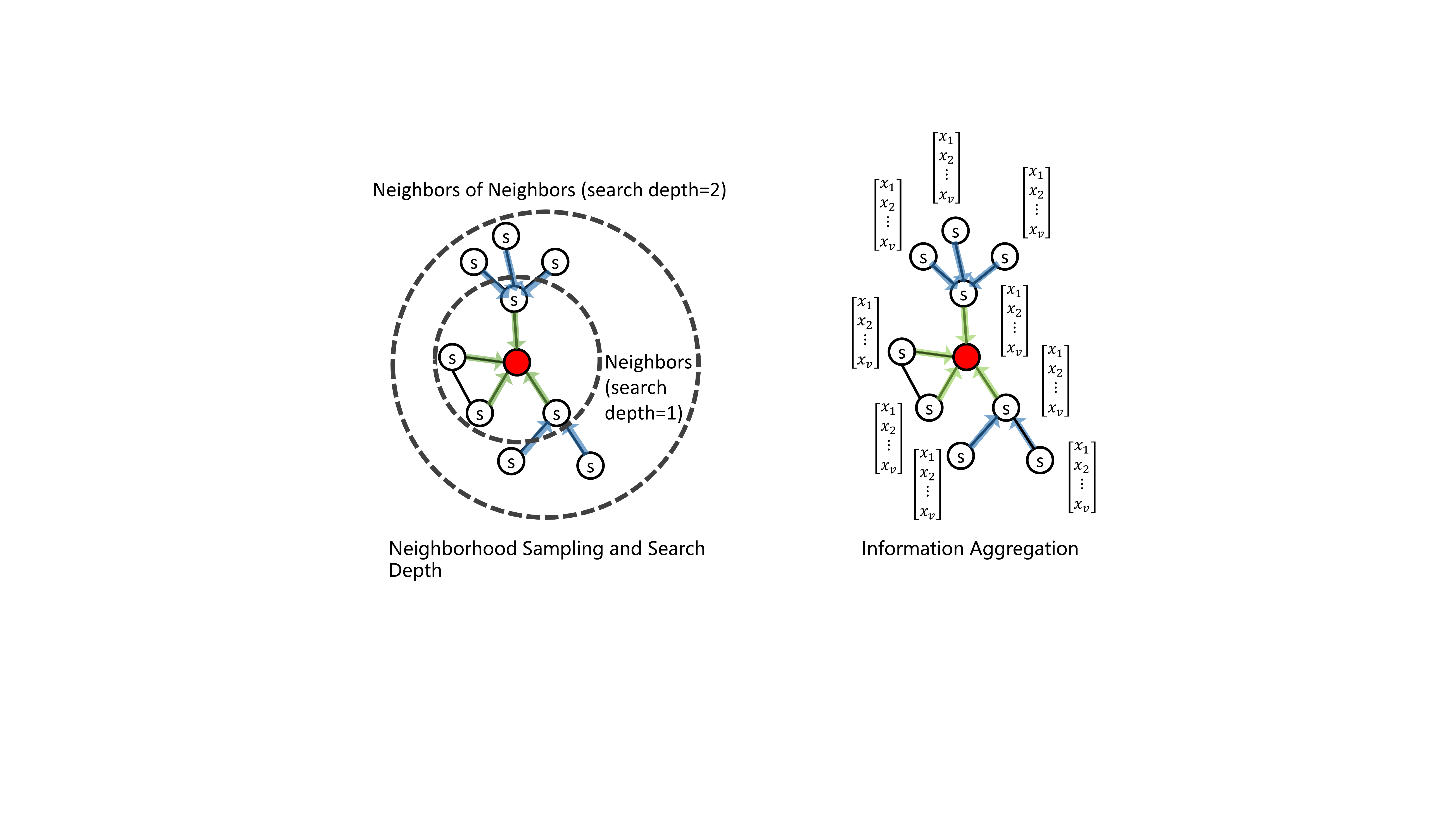}       
    \caption{Neighborhood Sampling and Information Aggregation of GraphSAGE}
    \label{fig:fig1}
\end{figure}

The message passing workflow of GraphSAGE mainly consists of two components, which is shown in Figure~\ref{fig:fig1}. 
The first component is neighbourhood sampling of the input graph. 
The second component is aggregating information at each search depth. 
Using this message passing mechanism, a GNN is able to embed into each node information about its neighbors and then employ the embedded information to make predictions. 
A major difference between GNN and other machine learning models is that GNN can deal with variable-sized graph inputs, while other models cannot. 
For a standard machine learning model, adding neighbor information as extra features works only for a graph with a fixed number of neighbors for each node. GNN can handle graphs whose nodes have a variable number of neighbors.

Once information is collected from neighboring nodes, a $\text{MEAN}_k$, $\forall k \in \{1, \dots, K\}$, aggregate function is used to calculate the average information. 
For each node, the GraphSAGE model iteratively aggregates information from the node’s neighborhood, the node’s neighbors’ neighborhood, and so on. 
During training, the aggregated information $h_{N(v)}^{k}$ of node $v$ at the $k$th layer can be expressed as  
\begin{equation}
h_{N(v)}^{k} = \text{MEAN}_k \left( \{ h_u^{(k-1)}, \ \forall u \in N(v) \} \right),
\label{eq:mean_aggregation}
\end{equation}
where $N(v)$ is the sampled neighborhood of node $v$ and $h_u^{(k-1)}$ represents the embedding of node $u$ in the previous layer. 
The aggregated embeddings of the sampled neighborhood $h_{N(v)}^{k}$ are then concatenated with the node’s embedding from the previous layer $h_v^{(k-1)}$. 
The embedding of node $v$ at layer $k$ can be calculated by applying the model’s trainable parameters $W^{k}$ and passing the result through a non-linear activation function $\sigma$ (e.g., ReLU):
\begin{equation}
h_v^{k} = \sigma \left( W^{k} \cdot \left[ h_v^{(k-1)} \ \| \ h_{N(v)}^{k} \right] \right),
\label{eq:node_update}
\end{equation}
where $\|$ denotes vector concatenation.  

The final embedding of node $v$ at the last layer $K$ is expressed as $z_v$:
\begin{equation}
z_v = h_v^{K}, \quad \forall v \in V.
\label{eq:final_embedding}
\end{equation}

\section{Case Study}\label{case-study}

\subsection{Data Description}

To demonstrate and evaluate the applicability of the proposed model, a case study was carried out using pavement condition inventory data (more than 110,000 data points each year) from Texas Department of Transportation (TxDOT). The data used was collected from pavement sections (around 0.5 mile in length) across Texas between 2014 to 2018. Each pavement section is labeled with a unique reference marker, which was used create the spatial relationships between a section and its neighbors in this research. If one section’s ending reference marker is the same as another section’s beginning reference marker, these two sections are considered connected and neighbors. 

The variables used in this study contains key attributes of pavement condition observations and other related variables as shown in Table \ref{tab:table1}. Although the initial condition right after a maintenance and rehabilitation treatment is usually used to model post-treatment deterioration, that information is not available in the dataset of this case study. For this reason, we didn’t create a variable indicating the post-treatment condition. Instead, we used annual inspected condition data where the average time between a treatment and the next inspection is around half year.

\begin{table}[htbp]
\centering
\caption{Variables used in the case study}
\label{tab:table1}
\renewcommand{\arraystretch}{1.2}
\begin{tabular}{p{0.28\linewidth} p{0.65\linewidth}}
\toprule
\textbf{Variable} & \textbf{Description} \\
\midrule
Pavement condition indicators & Twelve condition indicators were used in this case study. Detailed information about these indicators can be found in Table \ref{tab:table2}. \\
\addlinespace
Time since last treatment & The time difference (in years) between the last applied treatment and 2018. \\
\addlinespace
Traffic & The current 18-kip ESAL value for the data collection section. Values are stored in thousands. \\
\addlinespace
Road work records & Treatment types recorded by TxDOT. Detailed information about treatment types can be found in Table \ref{tab:table3}. \\
\addlinespace
Climatic regions & Climate zones based on temperature and precipitation. \\
\addlinespace
Type of pavement surface & ACP surface types categorized based on the similarity of characteristics. \\
\addlinespace
Functional class & In Texas, highways are categorized into seven groups based on their function, with each group containing sub-groups \citep{txdot2022glossary}. \\
\bottomrule
\end{tabular}
\end{table}

\subsubsection{Condition Indicator}
The condition variables include 12 flexible pavement condition indicators are shown in Table \ref{tab:table2}. The TxDOT pavement management system stores three scores that represent the general condition of a pavement \citep{xu2021development}. The Distress Score (DS) reflects the amount of visible surface deterioration of a pavement, with a range from 1 (the most distress) to 100 (the least distress). The Ride Score (RS) is a measure of the pavement’s roughness, ranging from 0.1 (the roughest) to 5.0 (the smoothest). The Condition Score (CS) represents the pavement’s overall condition in terms of both distress and ride quality ranging from 1 (the worst condition) to 100 (the best condition). Other indicators include shallow rutting, deep rutting, patching, failures, block cracking, alligator cracking, longitudinal cracking, transverse cracking, and international roughness index (IRI). 

\begin{table}[htbp]
\centering
\caption{Pavement condition indicators}
\label{tab:table2}
\renewcommand{\arraystretch}{1.15}
\begin{tabular}{>{\centering\arraybackslash}p{0.04\linewidth} p{0.40\linewidth} 
                >{\centering\arraybackslash}p{0.15\linewidth} 
                >{\centering\arraybackslash}p{0.20\linewidth}}
\toprule
\textbf{\#} & \textbf{Condition Indicator} & \textbf{Unit} & \textbf{Range} \\
\midrule
1  & Shallow rutting (0.25--0.49 inch depth) & Percentage & 0--100 \\
2  & Deep rutting (0.50--0.99 inch depth)    & Percentage & 0--100 \\
3  & Patching                                & Percentage & 0--100 \\
4  & Failures                                & Quantity   & $\geq 0$ \\
5  & Block cracking                          & Percentage & 0--100 \\
6  & Alligator cracking                      & Percentage & 0--100 \\
7  & Longitudinal cracking                   & Foot       & $\geq 0$ \\
8  & Transverse cracking                     & Quantity   & $\geq 0$ \\
9  & IRI                                     & inch/mile  & $\geq 0$ \\
10 & Ride score                              & N.A.       & 0--5 \\
11 & Distress score                          & N.A.       & 0--100 \\
12 & Condition score                         & N.A.       & 0--100 \\
\bottomrule
\end{tabular}
\end{table}

\subsubsection{Pavement Type}
\begin{figure}[htbp]
    \centering
    \includegraphics[width=0.75\linewidth]{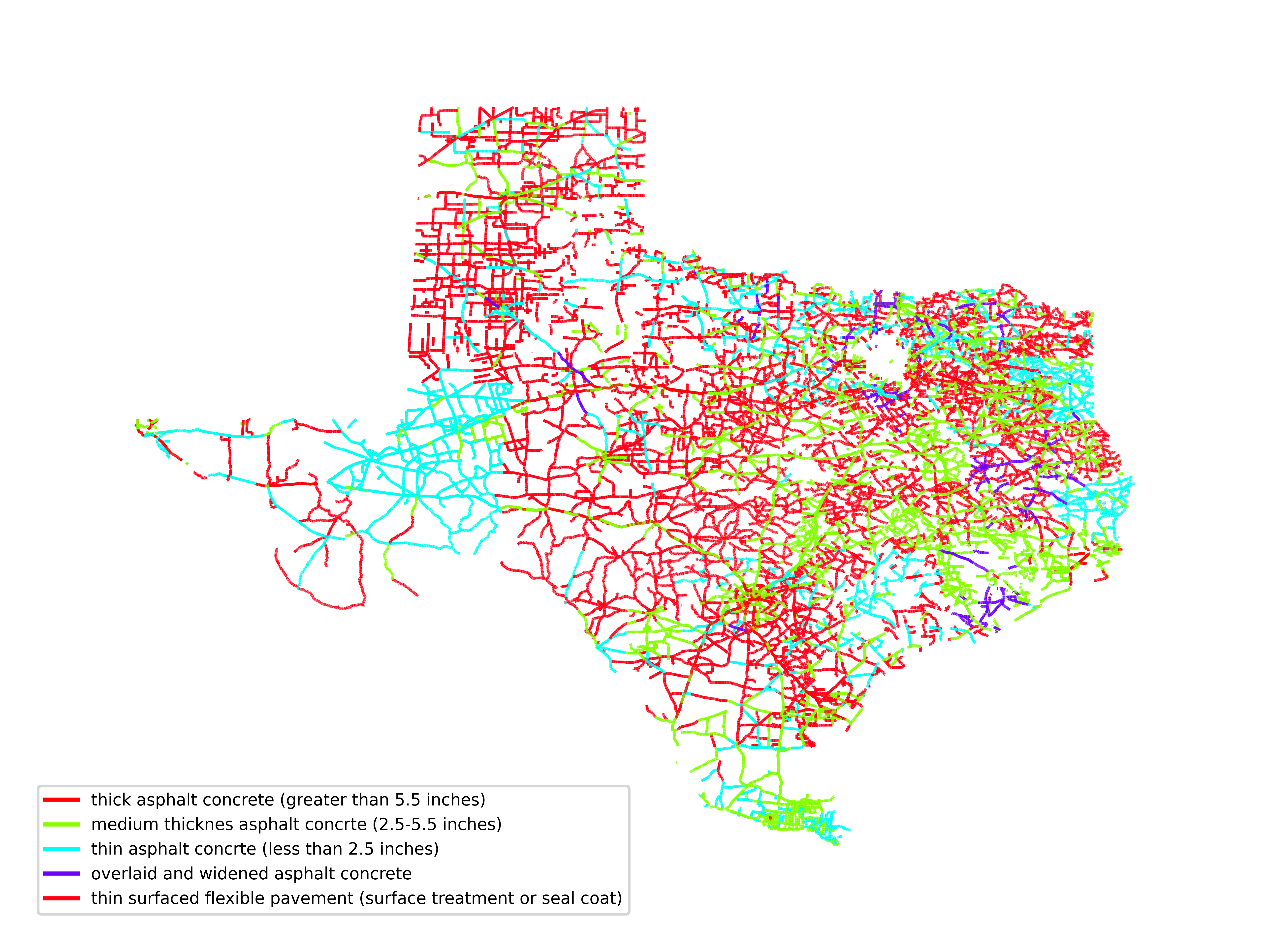}
    \caption{Map of Data Used in this Case Study by Different Pavement Types}
    \label{fig:fig2}
\end{figure}

There are ten different types of pavements in the TxDOT pavement management system. In this research, five asphalt pavement types (codes: 4, 5, 6, 9, and 10) were used Figure \ref{fig:fig2}. Code 4 represents thick asphalt concrete (greater than 5.5”). Code 5 indicates medium thickness asphalt concrete (2.5-5.5”). Code 6 represents thin asphalt concrete (less than 2.5”). Code 9 represents overlaid and widened asphalt concrete pavement. Code 10 represents thin surfaced flexible pavement (surface treatment or seal coat). 

\subsubsection{Functional Classification}
\begin{figure}[htbp]
    \centering
    \includegraphics[width=0.75\linewidth]{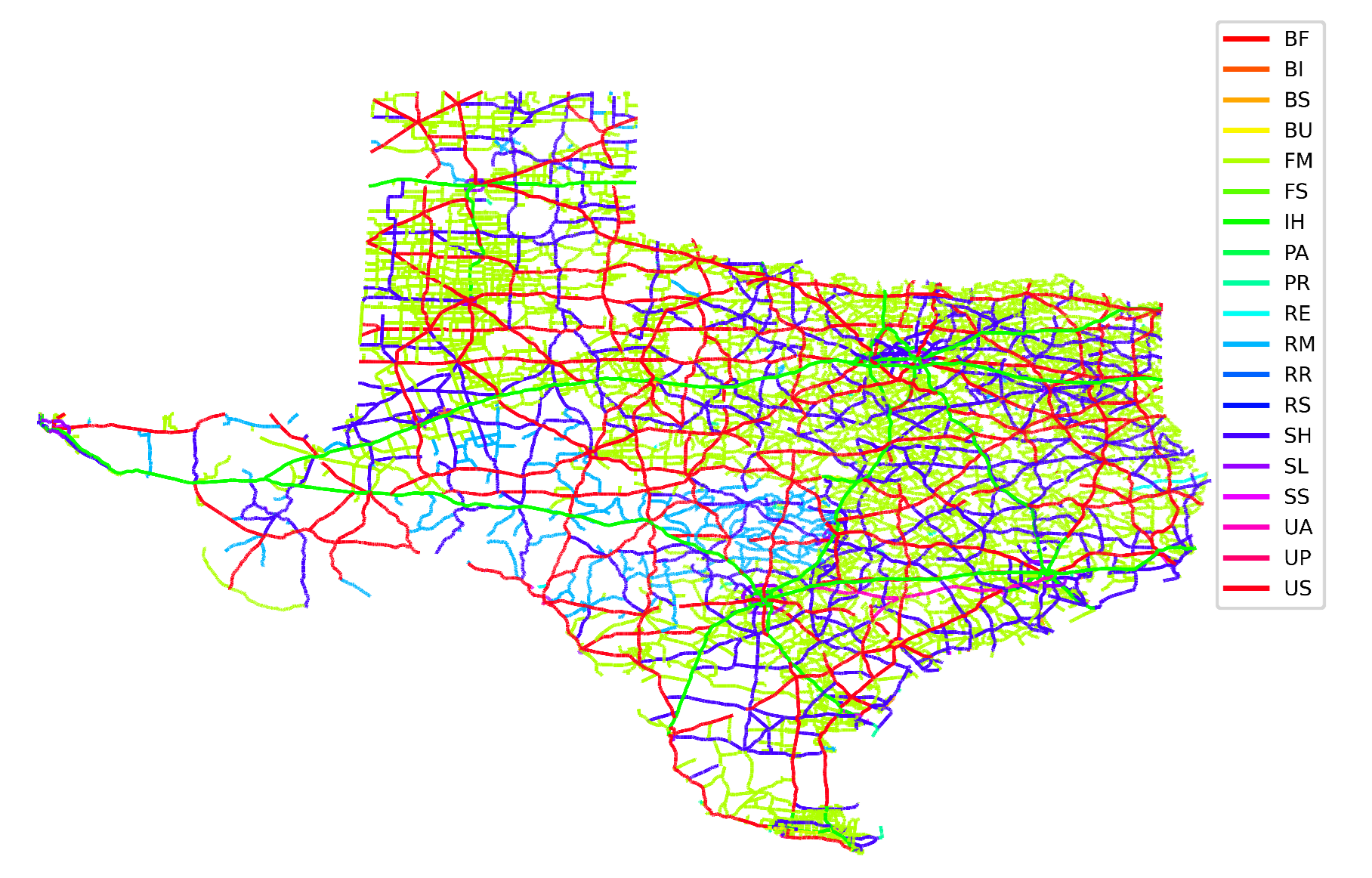}
    \caption{Map of Data Used in this Case Study by Different Functional Class}
    \label{fig:fig3}
\end{figure}

In Texas, highways are categorized into different groups based on their function \citep{xu2021development}. In this case study, 19 different groups of highways are used. As shown in Figure \ref{fig:fig3}, most of the highways fall into groups of Farm-to-Market (FM), State Highway (SH), US Highway (US), and Interstate Highway (IH), which corresponds to more than 90 percent of all the records. 

\subsubsection{Climate}
\begin{figure}[htbp]
    \centering
    \includegraphics[width=0.75\linewidth]{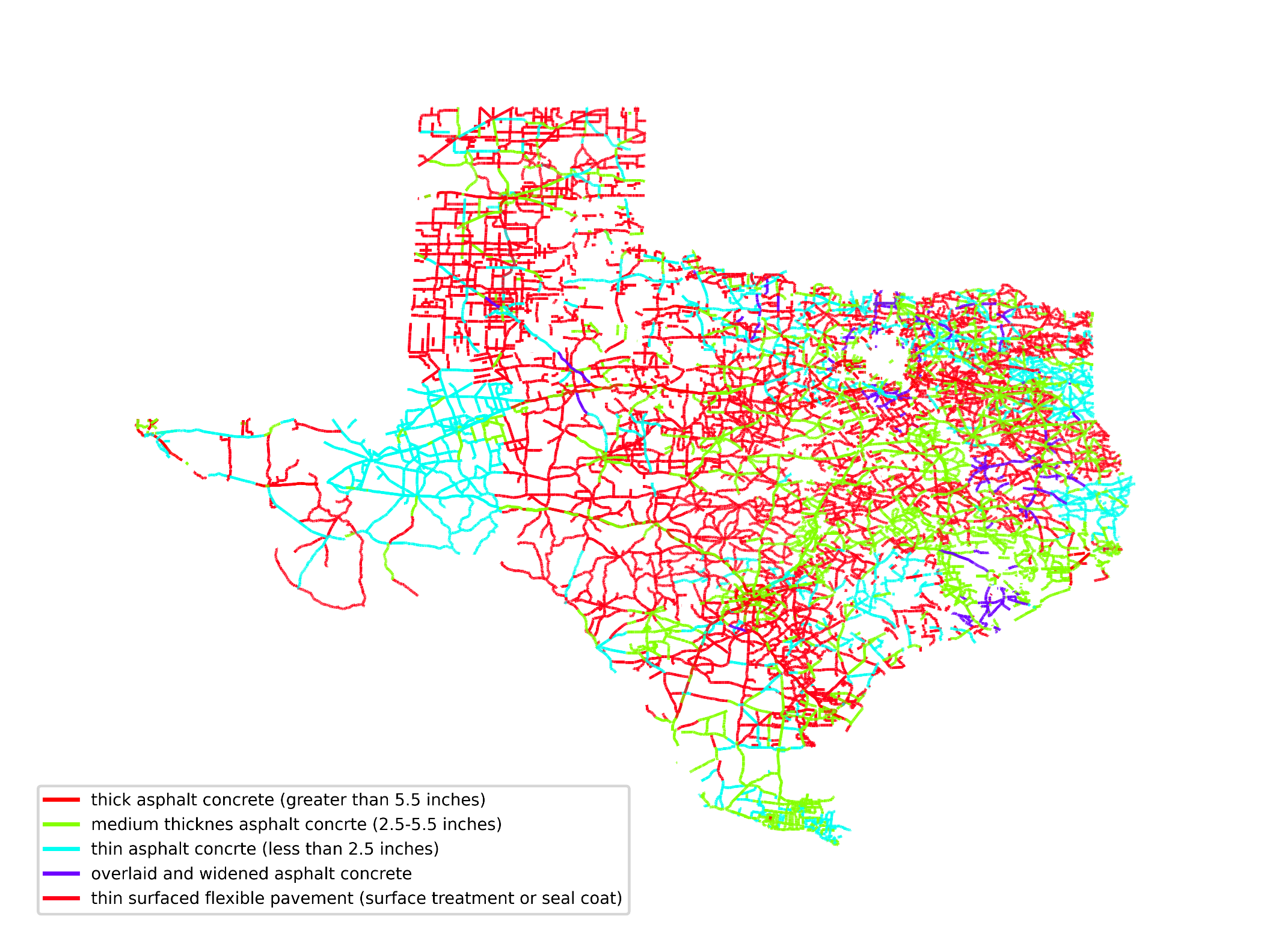}
    \caption{Climatic Zones with County Index}
    \label{fig:fig4}
\end{figure}

The climate information was obtained from the National Oceanic and Atmospheric Administration (NOAA) database. We used 30-year annual average temperature and precipitation as representative of the climate. For counties without a weather station, the average information from the adjacent counties was used. For counties with multiple weather stations, the average information from these stations was used. After acquiring the temperature and precipitation statistics for each county and consulting with TxDOT engineers, thresholds of 61.25 and 70.0 degrees (low, medium, high) for temperature and 16 and 38 inches (low, medium, high) for precipitation were used to group counties into different climate zones: west, east, north, south, and central regions (Figure \ref{fig:fig4}).

\subsubsection{Traffic}

\begin{figure}[htbp]
    \centering
    \includegraphics[width=0.75\linewidth]{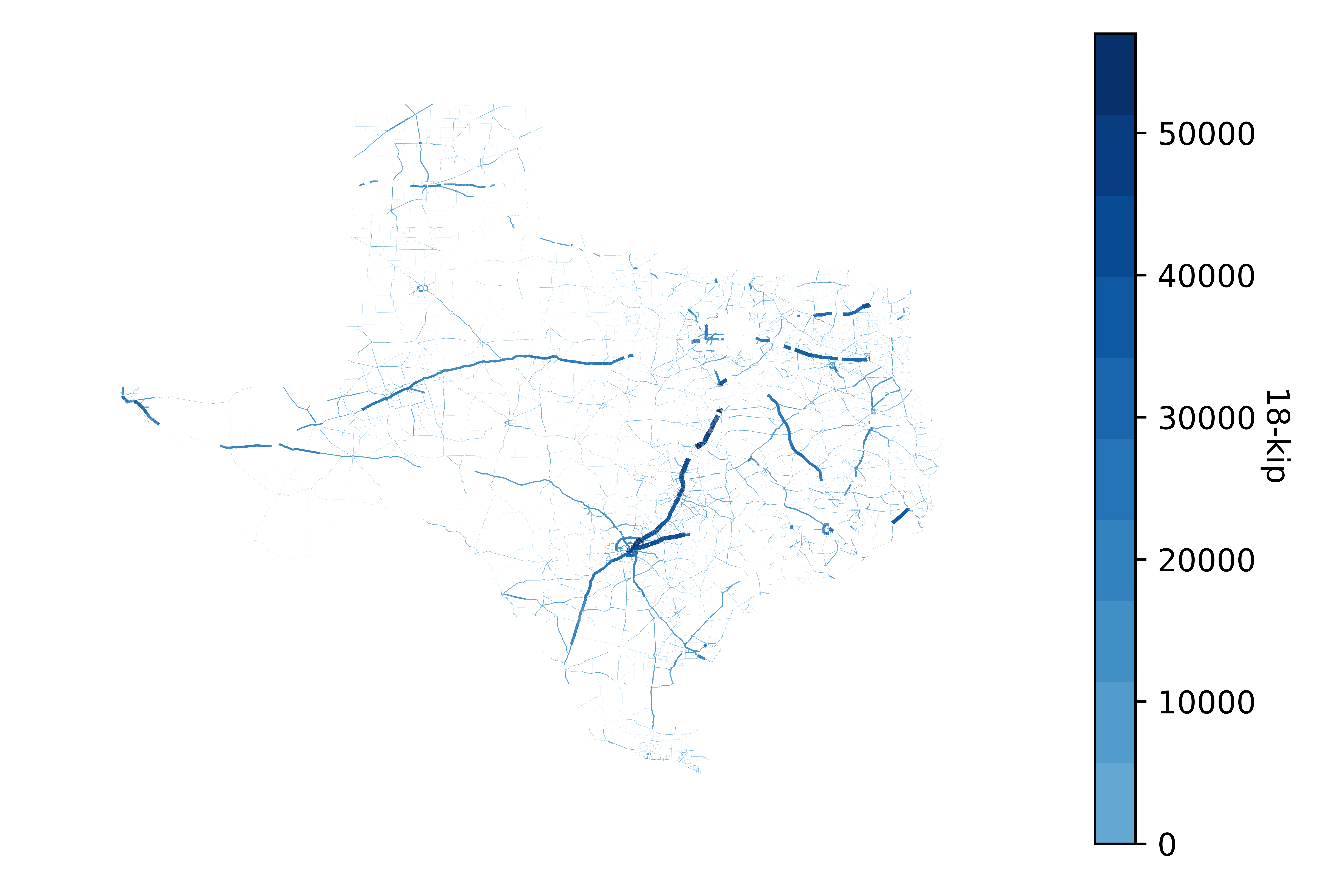}
    \caption{Map of Traffic Distribution }
    \label{fig:fig5}
\end{figure}

In this project, the 20-year projected ESALs were used to represent the traffic characteristic of each pavement section. The distribution of the traffic was plotted in Figure \ref{fig:fig5}. 

\subsubsection{Work History}
In this case study, we used pavement maintenance and rehabilitation history data collected by TxDOT. The maintenance dataset contains information about the type of treatments shown in Table \ref{tab:table3}, when they were implemented, and which pavement sections they were applied to. The work history records were converted to dummy variables representing each individual treatment when used in the modeling process.

\begin{table}[htbp]
\centering
\caption{TxDOT asphalt pavement treatment levels}
\label{tab:table3}
\begin{tabular}{p{0.30\linewidth} p{0.60\linewidth}}
\toprule
\textbf{Treatment Level} & \textbf{Treatment} \\
\midrule
\multirow{6}{*}{Preventive Maintenance (PM)} 
 & Cape Seal \\
 & Fog Seal \\
 & Micro-surfacing \\
 & Seal Coat \\
 & Thin Overlay (2'' thick or less) \\
 & Ultra-Thin Friction Course \\
\addlinespace
\multirow{5}{*}{Light Rehabilitation (LR)} 
 & Base Repair and Seal \\
 & Cold In-Place Recycling \\
 & Hot In-Place Recycling \\
 & Mill and Inlay; or Mill, Seal, and Thin Overlay \\
 & Overlay $>$ 2'' Thick but $<$ 3'' \\
\addlinespace
\multirow{5}{*}{Medium Rehabilitation (MR)} 
 & Base Repair, Spot Seal, Edge Repair, and Overlay \\
 & Level Up and Overlay \\
 & Mill and Overlay \\
 & Mill, Stabilize Base, and Seal \\
 & Overlay Between 3'' and 5'' \\
\addlinespace
\multirow{3}{*}{Heavy Rehabilitation (HR)} 
 & Full Depth Reclamation (Pulverization and Resurfacing) \\
 & Mill, Cement Stabilize Base, and Overlay \\
 & Reconstruction \\
 & Thick Overlay $>$ 5'' \\
\bottomrule
\end{tabular}
\end{table}

\subsection{Models}
In this case study, we developed performance models for each of the 12 condition indicators using the standard machine learning models: classification and regression trees (CART), neural network (NN), linear regression (LR) and the proposed GraphSAGE model. All machine learning models were implemented in scikit-learn \citep{buitinck2013api} and default hyperparameter values were used. GraphSAGE models were implemented in PyTorch Geometric (PyG), which is a Python library supporting many types of deep learning on graphs. PyG makes it easy to build a deep learning model though customizing predefined graph neural network layers \citep{fey2019fast}. We tuned the hyperparameters, number of layers and number of hidden channels per layer, by optimizing model performance. We varied one hyperparameter at a time in this tuning process. For the GraphSAGE model, we finally chose two layers (K=2) and the number of hidden channels of both layers were set to 256.

For each condition indicator’s deterioration model, the target variable is in its 2018 values and the features include 2014-2017 historical data of all condition indicators, maintenance work record, traffic, road functional class, climate zones, and pavement type information. We handle the inter-dependencies between condition indicators through including them into each other’s features. For example, when modeling IRI as the target, previous years’ cracking, rutting, and patching ratings were used as features. The selection of the features is based on data availability and the pavement deterioration models currently used by TxDOT \citep{xu2021development}. Historical condition data from 2001 to 2017 were evaluated and the results show that the effect of historical conditions reach its maximum around 4-5 years. Data beyond that time range has little impact on model performance. As a result, previous four years’ condition data (2014-2017) were included in the feature set. 20\% of the dataset is used for testing and the rest for training the models. R2-score, Mean Squared Error (MSE), and Mean absolute error (MAE) are used to measure and evaluate the performance of different models.

\subsection{Results}
The modeling results are provided in Table \ref{tab:table4}, which lists the r2-score, MSE, and MAE for each condition indicator. Negative R2-score indicates that the data cannot explain the target variable and the models perform poorly at predicting the testing set. It is probably because the target variables have little variation in the dataset. For each of the condition indicator, there are more than 110,000 data points (pavement sections). The number of data points used for each indicator were slightly different because of data availability. The proposed GraphSAGE model, which combines features from neighboring nodes, achieves better performance than the machine learning regression models using the scikit-learn library. The multivariate linear regression model, on average, has the best performance among the machine learning models. The worst performance is observed for the decision tree model. The best performances in terms of R2-score observed are for the IRI (0.87) and Ride Score (0.85), both of which are measurements of the pavement roughness and Ride Score is a linear transformation of IRI. The reason why roughness models have better results than other indicators is probably because the deterioration of roughness (i.e. changes between consecutive years) is more linear compared with other distress indicators. The r2-scores for other condition indicators are between 0.10 and 0.60. While the multivariate linear regression model gives the best results among all machine learning models, the improvement brought by GraphSAGE model ranges from 0\% to 20\% in R2-score.

\begin{table}[htbp]
\centering
\caption{Results of different models}
\label{tab:table4}
\setlength{\tabcolsep}{4pt}
\begin{tabular}{l *{4}{ccc}}
\toprule
\multirow{2}{*}{\textbf{Indicator}} 
 & \multicolumn{3}{c}{\textbf{LR}} 
 & \multicolumn{3}{c}{\textbf{CART}} 
 & \multicolumn{3}{c}{\textbf{NN}} 
 & \multicolumn{3}{c}{\textbf{GraphSAGE}} \\
\cmidrule(lr){2-4} \cmidrule(lr){5-7} \cmidrule(lr){8-10} \cmidrule(lr){11-13}
 & R$^2$ & MSE & MAE & R$^2$ & MSE & MAE & R$^2$ & MSE & MAE & R$^2$ & MSE & MAE \\
\midrule
Shallow Rutting      & 0.59 & 70.73  & 5.33 & 0.28  & 125.28 & 6.79 & 0.40  & 104.50 & 7.11 & 0.64 & 62.99  & 4.90 \\
Deep Rutting         & 0.47 & 6.51   & 1.16 & 0.00  & 12.42  & 1.39 & 0.44  & 6.84   & 1.29 & 0.51 & 6.00   & 1.04 \\
Patching             & 0.11 & 30.33  & 2.08 & -0.29 & 44.30  & 1.89 & -0.40 & 47.90  & 3.50 & 0.31 & 23.51  & 1.70 \\
Failures             & 0.08 & 0.44   & 0.15 & -1.10 & 1.02   & 0.15 & -0.24 & 0.61   & 0.33 & 0.12 & 0.43   & 0.12 \\
Block Cracking       & 0.08 & 6.43   & 0.66 & -0.58 & 6.43   & 0.55 & -56.55& 404.82 & 11.00& 0.13 & 6.11   & 0.49 \\
Alligator Cracking   & 0.17 & 16.59  & 1.71 & -0.37 & 27.68  & 1.74 & -0.42 & 28.72  & 2.87 & 0.35 & 13.07  & 1.43 \\
Longitudinal Cracking& 0.15 & 351.33 & 11.76& -0.18 & 489.46 & 12.27& 0.26  & 328.94 & 11.37& 0.30 & 287.10 & 10.25 \\
Transverse Cracking  & 0.16 & 0.41   & 0.30 & -0.28 & 0.64   & 0.27 & -0.46 & 0.73   & 0.49 & 0.22 & 0.39   & 0.27 \\
IRI                  & 0.84 & 296.03 & 10.56& 0.70  & 553.03 & 14.76& 0.85  & 276.10 & 10.25& 0.87 & 262.59 & 9.93 \\
Ride Score           & 0.85 & 0.07   & 0.17 & 0.70  & 0.14   & 0.24 & 0.08  & 0.46   & 0.45 & 0.85 & 0.07   & 0.17 \\
Distress Score       & 0.35 & 114.09 & 6.91 & -0.15 & 203.71 & 8.49 & 0.35  & 114.33 & 6.89 & 0.43 & 100.27 & 6.27 \\
Condition Score      & 0.42 & 133.03 & 7.43 & -0.03 & 241.69 & 9.26 & 0.42  & 134.92 & 7.92 & 0.50 & 115.77 & 6.78 \\
\bottomrule
\end{tabular}
\end{table}

\begin{figure}[htbp]
    \centering
    \includegraphics[width=0.75\linewidth]{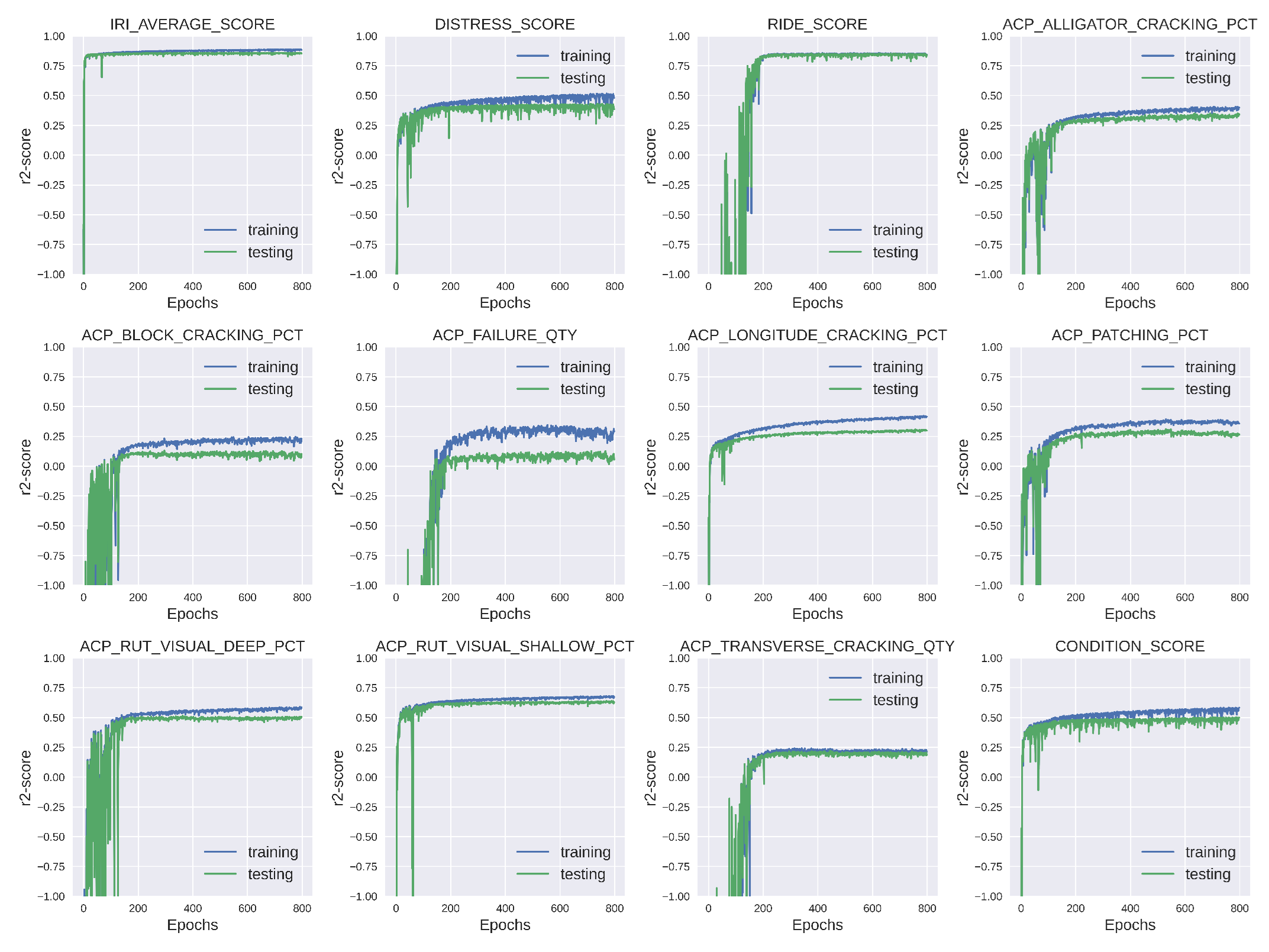}
    \caption{Training and Testing R2-score vs. Epoch of GraghSAGE Model}
    \label{fig:fig6}
\end{figure}

Figure \ref{fig:fig6} shows the training history of the GraphSAGE models. The modeling results are able to converge and achieve the highest r2-score after 100 to 800 epochs. The history for the training and testing dataset is labeled as training and the history for the testing dataset is labeled as testing. From the plots we can see that the models for longitudinal cracking and condition score could probably achieve higher value of R score if trained a little more epochs as the trend for r2-score on these datasets is still rising for the last few epochs. It can be found that the models for the indicators mentioned above has not been significantly over-learned in the training dataset, showing comparable performance on both training and testing datasets.

\section{Conclusions}
In this paper, we used graph neural network models to analyze pavement condition data for performance prediction. The pavement network is considered as a graph combining historical condition inventory data and spatial connections between neighboring sections. The spatial relationship between a section and its neighboring sections were then taken into account when developing the deterioration models. The results show that our model outperforms other machine learning models on a large-scale real-world network with more than 100,000 nodes and edges. The best R2-score result, 0.87, was obtained for IRI deterioration model.  The developed model can assist engineers and administrators in more effectively managing pavement assets through improved deterioration prediction. The results of this research can be integrated into existing pavement management system through two approaches. One is to use the trained model directly for condition prediction and the other is to consider neighboring sections’ information when developing new deterioration models. The pavement sections used in this research were spatially coupled by using the unique reference marker stored in TxDOT pavement management system. However, this approach can only recognize sections aligned linearly along the route. It does not recognize sections located nearby but not on the same route. Future research should include the latitude/longitude information of each section and consider sections within a certain radius. Future research can also address the inter-dependencies between condition indicators through combining all condition indicators into one multi-dimensional target.

\bibliographystyle{unsrtnat}
\bibliography{references}

\begin{thebibliography}{30}
\providecommand{\natexlab}[1]{#1}
\providecommand{\url}[1]{\texttt{#1}}
\expandafter\ifx\csname urlstyle\endcsname\relax
  \providecommand{\doi}[1]{doi: #1}\else
  \providecommand{\doi}{doi: \begingroup \urlstyle{rm}\Url}\fi

\bibitem[Zou and Madanat(2012)]{zou2012incorporating}
Bo~Zou and Samer Madanat.
\newblock Incorporating delay effects into airport runway pavement management systems.
\newblock \emph{Journal of Infrastructure Systems}, 18\penalty0 (3):\penalty0 183--193, 2012.

\bibitem[Jahanbakhsh et~al.(2016)Jahanbakhsh, Gao, and Zhang]{jahanbakhsh2016estimating}
Sam Jahanbakhsh, Lu~Gao, and Zhanmin Zhang.
\newblock Estimating spatial dependence associated with deterioration process of road network.
\newblock Technical report, 2016.

\bibitem[Atef and Moselhi(2014)]{atef2014modeling}
Ahmed Atef and Osama Moselhi.
\newblock Modeling spatial and functional interdependencies of civil infrastructure networks.
\newblock In \emph{Pipelines 2014: From Underground to the Forefront of Innovation and Sustainability}, pages 1558--1567. 2014.

\bibitem[Bernhardt and McNeil(2004)]{bernhardt2004agent}
Kristen L~Sanford Bernhardt and Sue McNeil.
\newblock An agent based approach to modeling the behavior of civil infrastructure systems.
\newblock In \emph{Engineering Systems Symposium, Tang Center, MIT}, 2004.

\bibitem[McDaniels et~al.(2007)McDaniels, Chang, Peterson, Mikawoz, and Reed]{mcdaniels2007empirical}
Timothy McDaniels, Stephanie Chang, Krista Peterson, Joey Mikawoz, and Dorothy Reed.
\newblock Empirical framework for characterizing infrastructure failure interdependencies.
\newblock \emph{Journal of Infrastructure Systems}, 13\penalty0 (3):\penalty0 175--184, 2007.

\bibitem[Rahman et~al.(2009)Rahman, Beznosov, and Mart{\'\i}]{rahman2009identification}
Hafiz~Abdur Rahman, Konstantin Beznosov, and Jos{\'e}~R Mart{\'\i}.
\newblock Identification of sources of failures and their propagation in critical infrastructures from 12 years of public failure reports.
\newblock \emph{International journal of critical infrastructures}, 5\penalty0 (3):\penalty0 220--244, 2009.

\bibitem[Panzieri et~al.(2004)Panzieri, Setola, and Ulivi]{panzieri2004agent}
S~Panzieri, R~Setola, and G~Ulivi.
\newblock An agent based simulator for critical interdependent infrastructures.
\newblock In \emph{Securing critical infrastructures, CRIS2004: conference on critical infrastructures}, pages 25--27. Grenoble FRANCE, 2004.

\bibitem[Dhatrak et~al.(2020)Dhatrak, Vemuri, and Gao]{dhatrak2020considering}
Omkar Dhatrak, Venkata Vemuri, and Lu~Gao.
\newblock Considering deterioration propagation in transportation infrastructure maintenance planning.
\newblock \emph{Journal of Traffic and Transportation Engineering (English Edition)}, 7\penalty0 (4):\penalty0 520--528, 2020.

\bibitem[Deng and Shi(2024)]{deng2024short}
Yong Deng and Xianming Shi.
\newblock Short-term predictions of asphalt pavement rutting using deep-learning models.
\newblock \emph{Journal of Transportation Engineering, Part B: Pavements}, 150\penalty0 (2):\penalty0 04024004, 2024.

\bibitem[Gao et~al.(2025)Gao, Din, Kim, and Senouci]{gao2025modeling}
Lu~Gao, Zia Din, Kinam Kim, and Ahmed Senouci.
\newblock Modeling the deterioration of pavement skid resistance and surface texture after preventive maintenance.
\newblock \emph{arXiv preprint arXiv:2507.01842}, 2025.

\bibitem[Alnaqbi et~al.(2024)Alnaqbi, Zeiada, and Al-Khateeb]{alnaqbi2024machine}
Ali Alnaqbi, Waleed Zeiada, and Ghazi~G Al-Khateeb.
\newblock Machine learning modeling of pavement performance and iri prediction in flexible pavement.
\newblock \emph{Innovative Infrastructure Solutions}, 9\penalty0 (10):\penalty0 385, 2024.

\bibitem[Peng et~al.(2025)Peng, Gao, Hong, and Sun]{peng2025evaluating}
Lidan Peng, Lu~Gao, Feng Hong, and Jingran Sun.
\newblock Evaluating pavement deterioration rates due to flooding events using explainable ai.
\newblock \emph{Buildings}, 15\penalty0 (9):\penalty0 1452, 2025.

\bibitem[Marcelino et~al.(2020)Marcelino, de~Lurdes~Antunes, Fortunato, and Gomes]{marcelino2020transfer}
Pedro Marcelino, Maria de~Lurdes~Antunes, Eduardo Fortunato, and Marta~Castilho Gomes.
\newblock Transfer learning for pavement performance prediction.
\newblock \emph{International Journal of Pavement Research and Technology}, 13\penalty0 (2):\penalty0 154--167, 2020.

\bibitem[Gao et~al.(2021{\natexlab{a}})Gao, Lu, and Ren]{gao2021deep}
Lu~Gao, Pan Lu, and Yihao Ren.
\newblock A deep learning approach for imbalanced crash data in predicting highway-rail grade crossings accidents.
\newblock \emph{Reliability Engineering \& System Safety}, 216:\penalty0 108019, 2021{\natexlab{a}}.

\bibitem[Hosseini et~al.(2020)Hosseini, Alhasan, and Smadi]{hosseini2020use}
Seyed~Amirhossein Hosseini, Ahmad Alhasan, and Omar Smadi.
\newblock Use of deep learning to study modeling deterioration of pavements a case study in iowa.
\newblock \emph{Infrastructures}, 5\penalty0 (11):\penalty0 95, 2020.

\bibitem[Yu and Gao(2023)]{yu2023pavement}
Ke~Yu and Lu~Gao.
\newblock Pavement missing condition data imputation through collective learning-based graph neural networks.
\newblock In \emph{International Conference on Transportation and Development 2023}, pages 416--423, 2023.

\bibitem[Garc{\'\i}a-Segura et~al.(2023)Garc{\'\i}a-Segura, Montalb{\'a}n-Domingo, Llopis-Castell{\'o}, Sanz-Benlloch, and Pellicer]{garcia2023integration}
Tatiana Garc{\'\i}a-Segura, Laura Montalb{\'a}n-Domingo, David Llopis-Castell{\'o}, Amalia Sanz-Benlloch, and Eugenio Pellicer.
\newblock Integration of deep learning techniques and sustainability-based concepts into an urban pavement management system.
\newblock \emph{Expert Systems with Applications}, 231:\penalty0 120851, 2023.

\bibitem[Gao et~al.(2022)Gao, Yu, and Lu]{gao2022missing}
Lu~Gao, Ke~Yu, and Pan Lu.
\newblock Missing pavement performance data imputation using graph neural networks.
\newblock \emph{Transportation research record}, 2676\penalty0 (12):\penalty0 409--419, 2022.

\bibitem[Haddad et~al.(2022)Haddad, Chehab, and Saad]{haddad2022use}
Angela~J Haddad, Ghassan~R Chehab, and George~A Saad.
\newblock The use of deep neural networks for developing generic pavement rutting predictive models.
\newblock \emph{International Journal of Pavement Engineering}, 23\penalty0 (12):\penalty0 4260--4276, 2022.

\bibitem[Gao et~al.(2021{\natexlab{b}})Gao, Yu, Hao~Ren, and Lu]{gao2021detection}
Lu~Gao, Yao Yu, Yi~Hao~Ren, and Pan Lu.
\newblock Detection of pavement maintenance treatments using deep-learning network.
\newblock \emph{Transportation Research Record}, 2675\penalty0 (9):\penalty0 1434--1443, 2021{\natexlab{b}}.

\bibitem[Lee et~al.(2019)Lee, Sun, and Lee]{lee2019development}
Yongjun Lee, Jongwan Sun, and Minjae Lee.
\newblock Development of deep learning based deterioration prediction model for the maintenance planning of highway pavement.
\newblock \emph{Korean Journal of Construction Engineering and Management}, 20\penalty0 (6):\penalty0 34--43, 2019.

\bibitem[Choi and Do(2019)]{choi2019development}
Seunghyun Choi and Myungsik Do.
\newblock Development of the road pavement deterioration model based on the deep learning method.
\newblock \emph{Electronics}, 9\penalty0 (1):\penalty0 3, 2019.

\bibitem[Gao et~al.(2023)Gao, Han, and Chen]{gao2023deep}
Lu~Gao, Zhe Han, and Yunshen Chen.
\newblock Deep learning--based pavement performance modeling using multiple distress indicators and road work history.
\newblock \emph{Journal of Transportation Engineering, Part B: Pavements}, 149\penalty0 (1):\penalty0 04022061, 2023.

\bibitem[Zhou et~al.(2021)Zhou, Okte, and Al-Qadi]{zhou2021predicting}
Qingwen Zhou, Egemen Okte, and Imad~L Al-Qadi.
\newblock Predicting pavement roughness using deep learning algorithms.
\newblock \emph{Transportation Research Record}, 2675\penalty0 (11):\penalty0 1062--1072, 2021.

\bibitem[Zhou et~al.(2020)Zhou, Cui, Hu, Zhang, Yang, Liu, Wang, Li, and Sun]{zhou2020graph}
Jie Zhou, Ganqu Cui, Shengding Hu, Zhengyan Zhang, Cheng Yang, Zhiyuan Liu, Lifeng Wang, Changcheng Li, and Maosong Sun.
\newblock Graph neural networks: A review of methods and applications.
\newblock \emph{AI open}, 1:\penalty0 57--81, 2020.

\bibitem[Hamilton et~al.(2017)Hamilton, Ying, and Leskovec]{hamilton2017inductive}
Will Hamilton, Zhitao Ying, and Jure Leskovec.
\newblock Inductive representation learning on large graphs.
\newblock \emph{Advances in neural information processing systems}, 30, 2017.

\bibitem[{Texas Department of Transportation}(2022)]{txdot2022glossary}
{Texas Department of Transportation}.
\newblock Highway designations glossary.
\newblock \url{https://www.txdot.gov/inside-txdot/division/transportation-planning/highway-designation/glossary.html}, 2022.
\newblock Accessed: July 5, 2022.

\bibitem[Xu et~al.(2021)Xu, Kim, Sabillon, Gao, Prozzi, et~al.]{xu2021development}
Hongbin Xu, Moo~Yeon Kim, Christian Sabillon, Lu~Gao, Jorge~A Prozzi, et~al.
\newblock Development of pavement performance models for pavement management incorporating treatment type.
\newblock Technical report, University of Texas at Austin. Center for Transportation Research, 2021.

\bibitem[Buitinck et~al.(2013)Buitinck, Louppe, Blondel, Pedregosa, Mueller, Grisel, Niculae, Prettenhofer, Gramfort, Grobler, et~al.]{buitinck2013api}
Lars Buitinck, Gilles Louppe, Mathieu Blondel, Fabian Pedregosa, Andreas Mueller, Olivier Grisel, Vlad Niculae, Peter Prettenhofer, Alexandre Gramfort, Jaques Grobler, et~al.
\newblock Api design for machine learning software: experiences from the scikit-learn project.
\newblock \emph{arXiv preprint arXiv:1309.0238}, 2013.

\bibitem[Fey and Lenssen(2019)]{fey2019fast}
Matthias Fey and Jan~Eric Lenssen.
\newblock Fast graph representation learning with pytorch geometric.
\newblock \emph{arXiv preprint arXiv:1903.02428}, 2019.

\end{thebibliography}

\end{document}